\documentclass[letterpaper, 10pt, journal, twoside]{IEEEtran}

\IEEEoverridecommandlockouts
\usepackage[pdftex]{graphicx} 

\usepackage[font=footnotesize]{caption}
\usepackage{amsmath,amsfonts}
\usepackage{dsfont} 

\usepackage{algorithm,algorithmic} 
\usepackage{url}

\usepackage[nobreak]{cite}

\usepackage[table]{xcolor}
\newcolumntype{C}[1]{>{\centering\arraybackslash}p{#1}}

\newcommand{\eg}{\textit{e.g.,}~} %
\newcommand{\ie}{\textit{i.e.,}~} %

\DeclareMathOperator*{\argmin}{arg\,min}

\newcommand{\U}       {U} 
\newcommand{\traj}    {\boldsymbol{\tau}} 

\usepackage[inline]{enumitem}\setlist[enumerate]*{label=(\roman*)}

\usepackage{siunitx}
\newcommand*{\sref}[1]{\S\ref{#1}}            
\newcommand*{\aref}[1]{Algorithm~\ref{#1}}  
\newcommand*{\tref}[1]{\text{TABLE~\ref{#1}}}   
\newcommand*{\fref}[1]{\text{Fig.~\ref{#1}}}  
\newcommand*{\eref}[1]{\text{Eq.~(\ref{#1})}}     

\title{ \huge
Leveraging Demonstrator-perceived Precision for Safe Interactive Imitation Learning of Clearance-limited Tasks
}

\author{
Hanbit Oh $^{\dagger}$ and Takamitsu Matsubara
\thanks{$^{\dagger}$ Corresponding author} 
\thanks{*This work is supported by NEDO, Grant Number [JPNP20006].}
\thanks{Both authors are with the Division of Information Science, Graduate School of Science and Technology, Nara Institute of Science and Technology, Japan}%
\thanks{© 2024 IEEE. Personal use of this material is permitted. Permission from IEEE must be obtained for all other uses, in any current or future media, including reprinting/republishing this material for advertising or promotional purposes, creating new collective works, for resale or redistribution to servers or lists, or reuse of any copyrighted component of this work in other works.}
\thanks{H. Oh and T. Matsubara, "Leveraging Demonstrator-perceived Precision for Safe Interactive Imitation Learning of Clearance-limited Tasks," in \textit{IEEE Robotics and Automation Letters (RA-L)}, 2024, doi: 10.1109/LRA.2024.3366755.}
}

\begin{document}

\maketitle
\begin{abstract}
Interactive imitation learning is an efficient, model-free method through which a robot can learn a task by repetitively iterating an execution of a learning policy and a data collection by querying human demonstrations. However, deploying unmatured policies for clearance-limited tasks, like industrial insertion, poses significant collision risks. For such tasks, a robot should detect the collision risks and request intervention by ceding control to a human when collisions are imminent. The former requires an accurate model of the environment, a need that significantly limits the scope of IIL applications. In contrast, humans implicitly demonstrate environmental precision by adjusting their behavior to avoid collisions when performing tasks. Inspired by human behavior, this paper presents a novel interactive learning method that uses \textit{demonstrator-perceived precision} as a criterion for human intervention called Demonstrator-perceived Precision-aware Interactive Imitation Learning (DPIIL). DPIIL captures precision by observing the speed-accuracy trade-off exhibited in human demonstrations and cedes control to a human to avoid collisions in states where high precision is estimated. DPIIL improves the safety of interactive policy learning and ensures efficiency without explicitly providing precise information of the environment. We assessed DPIIL's effectiveness through simulations and real-robot experiments that trained a UR5e 6-DOF robotic arm to perform assembly tasks. Our results significantly improved training safety, and our best performance compared favorably with other learning methods.
\end{abstract}

\section{Introduction}
\IEEEPARstart{I}{mitation} learning \cite{osa2018algorithmic} is an attractive way for robots to learn a policy for task automation by observing human demonstrations rather than manually engineering them using environmental models. Interactive Imitation Learning (IIL) \cite{celemin2022interactive} is a specific variant of this technique that optimizes a robot's policy by repeating the interactions between a robot that has executed its unmatured policy and a human who provides corrective demonstrations while observing their execution. Although standard imitation learning cannot determine how many human demonstrations are needed to ensure that a policy is learned, IIL allows a human to observe the execution of a policy being learned, making it possible to train until its performance is guaranteed more efficiently.

However, as in IIL, deploying unmatured policies poses significant collision risks in clearance-limited tasks, such as aperture-passing and ring-threading. To ensure safety, a robot must be aware of the risks of collisions and request intervention by ceding control to a human to avoid them. Detecting collision risks requires precision information, such as the narrowness of the environment, which provides the spatial context of collisions. Although precision can be obtained with a model of the environment, IIL is model-free, a situation that limits its applicability.

\begin{figure}[t]
    \centering
    \includegraphics[width=1.0\hsize]{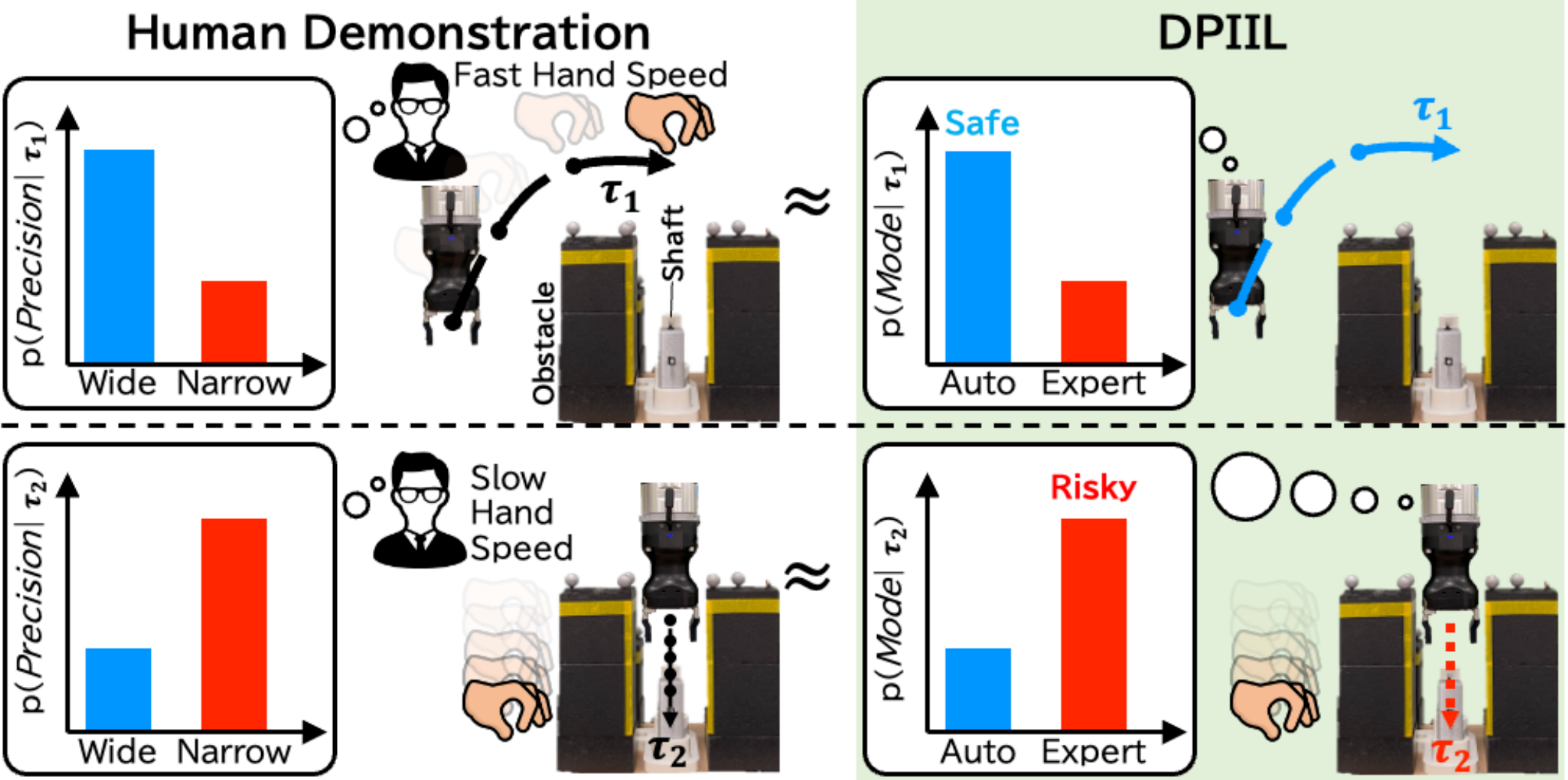}
    \caption{
        Overview of Demonstrator-perceived Precision-aware Interactive Imitation Learning (DPIIL).
        In clearance-limited tasks, demonstrator-perceived precision is in the mind of humans. By capturing this precision level from demonstration data and incorporating it into IIL, a robot can cede control to a human (expert mode, bottom) in high-precision areas while executing its policy (auto mode, top) in low-precision areas, thus enhancing safety.
}
    \label{fig:intro:punch}
\end{figure}

This study aims to develop an approach for safely applying IIL in clearance-limited tasks. To achieve this, the key idea is to identify environmental precision from human demonstrations in a model-free manner based on findings from behavioral psychology. We assume that humans can perceive environmental precision based on their understanding of the environment, and such precision can be captured from the \textit{speed-accuracy trade-off} \cite{wickelgren1977speed} exhibited by humans during task execution. For example, in a task that reaches a shaft between obstacles, humans slow down to increase their accuracy as the gap between obstacles narrowers (\fref{fig:intro:punch}, bottom). As such, \textit{demonstrator-perceived precision} can be captured from human movement speed and is used to estimate the collision risk of IIL in clearance-limited tasks.

Therefore, this paper presents a novel interactive learning approach that incorporates demonstrator-perceived precision as intervention criteria (\fref{fig:intro:punch}): Demonstrator-perceived Precision-aware Interactive Imitation Learning (DPIIL). By employing a leader-following teleoperation system where a robot directly follows human hand movements, the human's speed-accuracy trade-off is directly reflected in demonstrations. This allows the speed of a robot controlled by a human to reflect the demonstrator-perceived precision. We introduce a precision estimator that learns to capture such speed distribution from demonstrations and approximates the precision for given states. Since DPIIL solicits human intervention in states where the estimated precision must be extremely high (\ie risk of collisions is excessive), 
DPIIL enhances the safety of IIL in clearance-limited tasks.

To summarize, the key contributions of this paper are as follows:
\begin{enumerate*}
    \item We develop a novel method to estimate collision risk associated with environmental precision by leveraging demonstrator-perceived precision.
    \item We present a safe IIL algorithm, DPIIL (Fig. 1), which uses collision risk as criteria to request human interventions when significant risk is estimated, inspired by risk-aware interactive design in previous studies \cite{menda2019ensembledagger, zhang2016safedagger, Hoque2021lazydagger, hoque2021thriftydagger}.
    \item We validate our method (DPIIL) in clearance-limited simulations (\eg aperture-passing and ring-threading tasks) and in real-robot experiments (\eg shaft-reaching and ring-threading tasks). The results show significantly improved training phase safety compared to other learning methods.
\end{enumerate*}

\section{Related Work}
\subsection{Interactive Imitation Learning}
Interactive Imitation Learning (IIL) \cite{celemin2022interactive} allows robots to acquire optimal actions through human-robot online interaction, such as injecting disturbances into human demonstrations \cite{laskey2017dart, oh2023bdi} or requesting human interventions for corrections during task execution \cite{ross2011dagger}. The latter has become a prominent method of efficient imitation learning since it provides theoretical performance guarantees \cite{ross2011dagger} without continuous human guidance, as required by the former. However, in tasks with limited clearances, such a standard IIL is impractical, since executing the robot's unmatured policies can lead to significant collisions.

\subsection{Risk-aware Interactive Imitation Learning}
Several risk-aware approaches in IIL have been studied to improve safety by estimating the risk of robot executions in given states and requesting human intervention when encountering risky states. One approach to guarantee IIL safety is using the risk awareness of humans based on their understanding of the environment. Humans continuously monitor a robot's execution and intervene when a robot encounters risky states and provide corrective action \cite{kelly2019hgdagger} or reset the task execution \cite{hoque2023fleet}. However, these approaches have the constraint of forcing users to constantly monitor the robot.

Instead of continuous human monitoring, other approaches have been investigated that leverage robotic risk awareness based on their policy analysis \cite{ menda2019ensembledagger,zhang2016safedagger, hoque2021thriftydagger,Hoque2021lazydagger}. These approaches allow a robot to quantify the execution risks and actively ask a human to intervene when the risk exceeds a threshold. Previous research defined risk indicators as the uncertainty of a robot's decision about the visited state \cite{menda2019ensembledagger} or the discrepancy between the actions proposed by a robot's policy and a human expert \cite{zhang2016safedagger}. However, neither metric can detect collision risks since a robot still lacks precision information of its environment. Although Hoque et al. introduced a precision estimation metric \cite{hoque2021thriftydagger}, it requires a robot to experience hundreds of collisions by itself to optimize the precision estimator; thus, this metric is limited in practical application. Alternatively, this paper explores implicitly estimating environmental precision from human demonstrations without requiring collision experiences.

\subsection{Speed-accuracy Trade-off in Clearance-limited Tasks}
During human demonstrations of clearance-limited robotic tasks, people carefully regulate a robot's speed through constrained spaces (\eg obstacles) to avoid collisions based on their understanding of the tasks and the environment \cite{nagengast2011risk}. This phenomenon has been extensively examined in neuroscience to identify the human balance between speed and accuracy, commonly called the speed-accuracy trade-off \cite{wickelgren1977speed}. This idea has also been studied in robotics to efficiently complete tasks while ensuring collision avoidance. In a path-planning context, speed-accuracy cost-maps have been achieved by providing explicit environmental dynamics models \cite{lin2009speed} and incorporating a heuristic search algorithm \cite{murphy2011risky}. Another approach uses human behavior as a guide without assuming an environmental dynamics model. Our previous IIL study \cite{oh2023bdi} explored this idea and exploited how humans factor in collision risk to regulate disturbances, which are injected into human demonstrations for policy robustification while ensuring feasibility. This paper delves deeper into this concept, proposing a novel approach that uses human behavior to capture the demonstrator-perceived precision to decide whether to request human intervention to mitigate the risk of collisions in another IIL framework.

\section{Problem Statement}
This section discusses a scenario where a human expert trains a robot to automatically perform a task. The system is designed to enable the robot to request human intervention when it needs help while executing a task. Then, a human expert takes control of the robot and guides it optimally only as long as the intervention is requested. These concepts are formulated based on previous research of imitation learning.

An environmental dynamics model is denoted as a Markovian with states $\mathbf{s}_t \in \mathcal{S}$, actions $\mathbf{a}_t \in \mathcal{A}$ and time horizon $T$. 
Parametric policy $\pi_{\theta}: \mathcal{S} \rightarrow \mathcal{A}$ is defined to control robot with parameter $\theta$. The human expert has a policy $\pi_{\theta^{*}}$ deciding optimal action $\mathbf{a}^*_t$ from state $\mathbf{s}_t$.
The goal of IL is to learn policy parameters $\theta^{L}$ that match the human expert's $\theta^{*}$ by minimizing a surrogate loss function $J$ as follows:
\begin{equation}
\min J(\theta^{L}) 
= \sum_{t=1}^{T} \mathbb{E}_{\mathbf{a}^*_t, \mathbf{s}_t\sim \traj^{*}_t}\left[\left\|\pi_{\theta^{L}}(\mathbf{s}_t)-\mathbf{a}^*_t\right\|_{2}^{2}\right],
\label{eq:il:obj}
\end{equation}
where $\traj^{*}_t$ is the trajectory distribution which is state-action pairs induced by expert's policy $\pi_{\theta^{*}}$ at step $t$.

Furthermore, a key aspect of IIL is to allow robots to request human intervention.
As such, a binary decision function $g(\mathbf{s}_t)=\mathds{1}$ is defined to determine whether a robot operates in an \textit{auto mode} ($g(\mathbf{s}_t)=0$, controlled by $\pi_{\theta^{L}}$) or an \textit{expert mode} ($g(\mathbf{s}_t)=1$, controlled by $\pi_{\theta^*}$).
Then cost $C$ of IIL is the total number of actions provided by a human expert along entire interactions. Thus, as in prior works \cite{zhang2016safedagger,menda2019ensembledagger,Hoque2021lazydagger, hoque2021thriftydagger}, the designed IIL aims to optimize policy while minimizing $C$.

Although policy optimization convergence is theoretically guaranteed \cite{ross2011dagger} while reducing human costs $C$ \cite{zhang2016safedagger,menda2019ensembledagger, Hoque2021lazydagger, hoque2021thriftydagger}, no risk awareness has been ensured. This situation is especially problematic in clearance-limited tasks, because the risk of collisions can greatly hinder task performance and significantly damage robots. The next section presents our novel IIL to estimate the environmental precision and prompt human intervention for collision risk mitigation.

\section{Demonstrator-perceived Precision-aware IIL}
In this section, we propose a novel Demonstrator-perceived Precision-aware Interactive Imitation Learning (DPIIL) algorithm that introduces a collision-risk-estimation metric based on demonstrator-perceived precision to increase safety during interactive policy learning.
In the following, \sref{sec:IV:A} describes how the demonstrator-perceived precision is estimated based on the speed-accuracy trade-off exhibited by humans, \sref{sec:proposal:collision} introduces the collision-risk-estimation metric from both the precision and the uncertainty analysis of a learned policy, \sref{sec:proposal:interaction} introduces an interaction design for mitigating collision risks, and \sref{sec:proposal:overview} describes DPIIL's overall algorithmic procedure. 

\subsection{Demonstrator-perceived Precision Estimation}\label{sec:IV:A}
First, we defined speed transformation function $f_v$, which computes speed $v_t$ from a pair of states along 1-step transitions:
$f_v(\mathbf{s}_{t},\mathbf{s}_{t+1})=v_t$. For the following formulation, $v_t$ is given by $f_v$.
Under this speed definition, human speeds are corrupted by state-dependent noise \cite{harris1998signal}, whose variance increases with the size of the input actions during demonstrations.
Such variations in demonstrations are called \textit{aleatoric uncertainty}, and a natural way to capture this uncertainty is to use a probabilistic neural network regression model \cite{nix1994aleatoric} that consists of two neural networks predicting the mean and variance (\ie aleatoric uncertainty), respectively.
Specifically, the speed estimator is defined as ${V}_{\lambda}(v_t|\mathbf{s}_t)$, which outputs the Gaussian distribution with mean network $\mu_{\lambda}(\mathbf{s}_t)$ and variance network $\sigma^2_{\lambda}(\mathbf{s}_t)$ for a given state $\mathbf{s}_t$ with parameter $\lambda$:
\begin{align}
    {V}_{\lambda}(v_t|\mathbf{s}_t) = \mathcal{N}(v_t |\mu_{\lambda}(\mathbf{s}_t),\sigma^2_{\lambda}(\mathbf{s}_t)).
\label{eq:pros:speed:estimator}
\end{align}

In practice, training dataset $\mathcal{D}$ for involving human speeds $v^*_t$ can be calculated by $f_v$ using transition $(\mathbf{s}_t, \mathbf{a}^*_t, \mathbf{s}_{t+1})$ of a human expert's trajectory: $\mathcal{D}=\{\mathbf{a}^*_t, \mathbf{s}_t, v^*_t\}^{T}_{t=1}$. For learning probabilistic speed estimator ${V}_{\lambda}(v_t|\mathbf{s}_t)$ in an imitation learning context, negative log-likelihood loss $L$ of the estimator is defined:
\begin{align}
    L(V_{\lambda}|\mathcal{D})= \sum^{T}_{t=1}-\log\mathcal{N}(v^*_t |\mu_{\lambda}(\mathbf{s}_t),\sigma^2_{\lambda}(\mathbf{s}_t)).
\label{eq:pros:speed:loss}
\end{align}
Therefore, the speed estimator's parameter $\lambda$ is optimized by minimizing the expected loss along the training dataset:
\begin{align}
\lambda^{\prime}=\argmin_{\lambda}
\mathbb{E}_{\mathcal{D}\sim \traj^{*}_t}[L(V_{\lambda}|\mathcal{D})].
\label{eq:pros:speed:obj}
\end{align}

Due to the speed-accuracy trade-off of humans \cite{wickelgren1977speed}, in narrow areas, the human speed mean and variance are decreased. For this human behavior, there are two types of modeling possibilities for precision estimator $\text{Pre}_{\lambda^{\prime}}(\mathbf{s}_t)$:
\begin{itemize}
    \item $\text{Pre}^{\mu}_{\lambda^{\prime}}(\mathbf{s}_t) = \{\mu_{\lambda^{\prime}}(\mathbf{s}_t)\}^{-1}$, where the precision is inversely proportional to the estimated speed's mean;
    \item $\text{Pre}^{\text{UCB}}_{\lambda^{\prime}}(\mathbf{s}_t) = \{\mu_{\lambda^{\prime}}(\mathbf{s}_t) + \sigma_{\lambda^{\prime}}(\mathbf{s}_t)\}^{-1}$, where the precision is inversely proportional to the estimated speed's Upper Confidence Bound (UCB), which is the sum of the mean and the standard deviation.
\end{itemize}
Implementing the former type is simpler, although it is expected to be less sensitive for capturing demonstrator-perceived precision than the latter type, which consider speed variance simultaneously. The DPIILs used for each precision model are defined as $\text{DPIIL}_{\mu}$ and $\text{DPIIL}_{\text{UCB}}$. 

\subsection{Collision Risk Estimation}
\label{sec:proposal:collision}
To estimate the collision risk, the robot must analyze not only the environment's precision but also the uncertainty of the learned policy for performing the task. Such policy uncertainty, called \textit{epistemic uncertainty}, stems from a lack of demonstration data and increases the risk that the robot will make unmatured decisions, which may induce collisions.

To capture the epistemic uncertainty of a learned policy, we employ an ensemble neural network as a policy model similar to the prior study \cite{menda2019ensembledagger}. 
As such, each component of the ensemble policies is learned by \eref{eq:il:obj}.
Then the ensemble of learned policies outputs actions for any given $\mathbf{s}_t$, and variances $\sigma_{\theta^{L}}^2(\mathbf{s}_t)$ of these actions can be interpreted as the level of epistemic uncertainty in the decision. Finally, to quantify the collision risk by comprehensively evaluating $\mathbf{s}_t$ regarding both precision $\text{Pre}_{\lambda^{\prime}}(\mathbf{s}_t)$ and the uncertainty of learned policy $\sigma_{\theta^{L}}^2(\mathbf{s}_t)$, collision risk $\text{Risk}(\mathbf{s}_t)$ is defined as:
\begin{equation}
    \text{Risk}(\mathbf{s}_t) = \text{Pre}_{\lambda^{\prime}}(\mathbf{s}_t)\cdot\sigma_{\theta^{L}}^2(\mathbf{s}_t).
\label{eq:pros:risk}
\end{equation}
Note that the efficacy of multiplying these two factors is: 
\begin{enumerate*}
    \item
    in open areas, precision $\text{Pre}_{\lambda^{\prime}}(\mathbf{s}_t)$ decreases, allowing for higher policy uncertainty $\sigma_{\theta^{L}}^2(\mathbf{s}_t)$ (\ie less demonstration data), and 
    \item
    in narrow areas, precision $\text{Pre}_{\lambda^{\prime}}(\mathbf{s}_t)$ increases, requiring lower policy uncertainty $\sigma_{\theta^{L}}^2(\mathbf{s}_t)$ (\ie more demonstration data). 
    \item
    Finally, once enough data has been collected to meet the appropriate policy uncertainty $\sigma_{\theta^{L}}^2(\mathbf{s}_t)$ for precision $\text{Pre}_{\lambda^{\prime}}(\mathbf{s}_t)$, the robot will request no more human intervention.
\end{enumerate*}

\subsection{Interaction Design}\label{sec:proposal:interaction}
Interaction using the collision risk estimation of \sref{sec:proposal:collision} is introduced to improve the safety of the interactive policy learning. To prompt human intervention triggered by collision risk, decision function $g(\mathbf{s}_t;\chi)$ is defined that is activated when $\textrm{Risk}(\mathbf{s}_t)$ exceeds threshold $\chi$:
\begin{align}\label{eq:RG-IIL:g}
g(\mathbf{s}_t;\chi) 
&= 
\begin{cases}
1, & \text{if } \textrm{Risk}(\mathbf{s}_t) > \chi\\
0, & \text{otherwise } \\
\end{cases},
\end{align}
which indicates whether state $\mathbf{s}_t$ is safe ($g(\mathbf{s}_t;\chi) = 0$) or risky ($g(\mathbf{s}_t;\chi) = 1$) regarding collisions. 
During a robot's training phase (\fref{fig:overview}-top), this decision function allows a robot to request human intervention (\ie expert mode) only in risky state $\mathbf{s}_t$ while deploying a learned policy (\ie auto mode) during the others.

\subsection{ DPIIL Overview}\label{sec:proposal:overview}

\begin{algorithm}[tb]
\caption{DPIIL}
\label{algorithm:DPIIL}
\begin{algorithmic}[1]
\small
\renewcommand{\algorithmicrequire}{\textbf{Input:}}
\renewcommand{\algorithmicensure}{\textbf{Output:}}
\REQUIRE Number of iterations $K$, threshold $\chi$
\ENSURE Parameter of learned policy $\theta^{L}_{K}$, parameter of precision estimator $\lambda_{K}$
\STATE Get the initial dataset through a human expert:
\\$\mathcal{D} = \{\mathbf{a}^*_t, v^*_t,\mathbf{s}_t\}_{t=1}^{T}$
\STATE Initialize ${\theta^{L}_0}$ and ${\lambda_0}$ by \eref{eq:il:obj} and \eref{eq:pros:speed:obj} on $\mathcal{D}$
\FOR {$k = 1$ to $K$}
    \STATE Get the dataset through interactions:
    \\ $\mathcal{D}_{k} = \{\mathbf{a}^*_t, v^*_t,\mathbf{s}_t\mid g(\mathbf{s}_t, \chi)=1\}_{t=1}^{T}$
    \STATE Aggregate datasets:
    $\mathcal{D} \gets \mathcal{D}\cup \mathcal{D}_{k}$ 
    \STATE Learn $\theta^{L}_{k}$ and ${\lambda_k}$ by \eref{eq:il:obj} and \eref{eq:pros:speed:obj} on $\mathcal{D}$
\ENDFOR
\end{algorithmic} 
\end{algorithm}

\begin{figure}[t!]
    \centering
    \includegraphics[width=0.8\hsize]{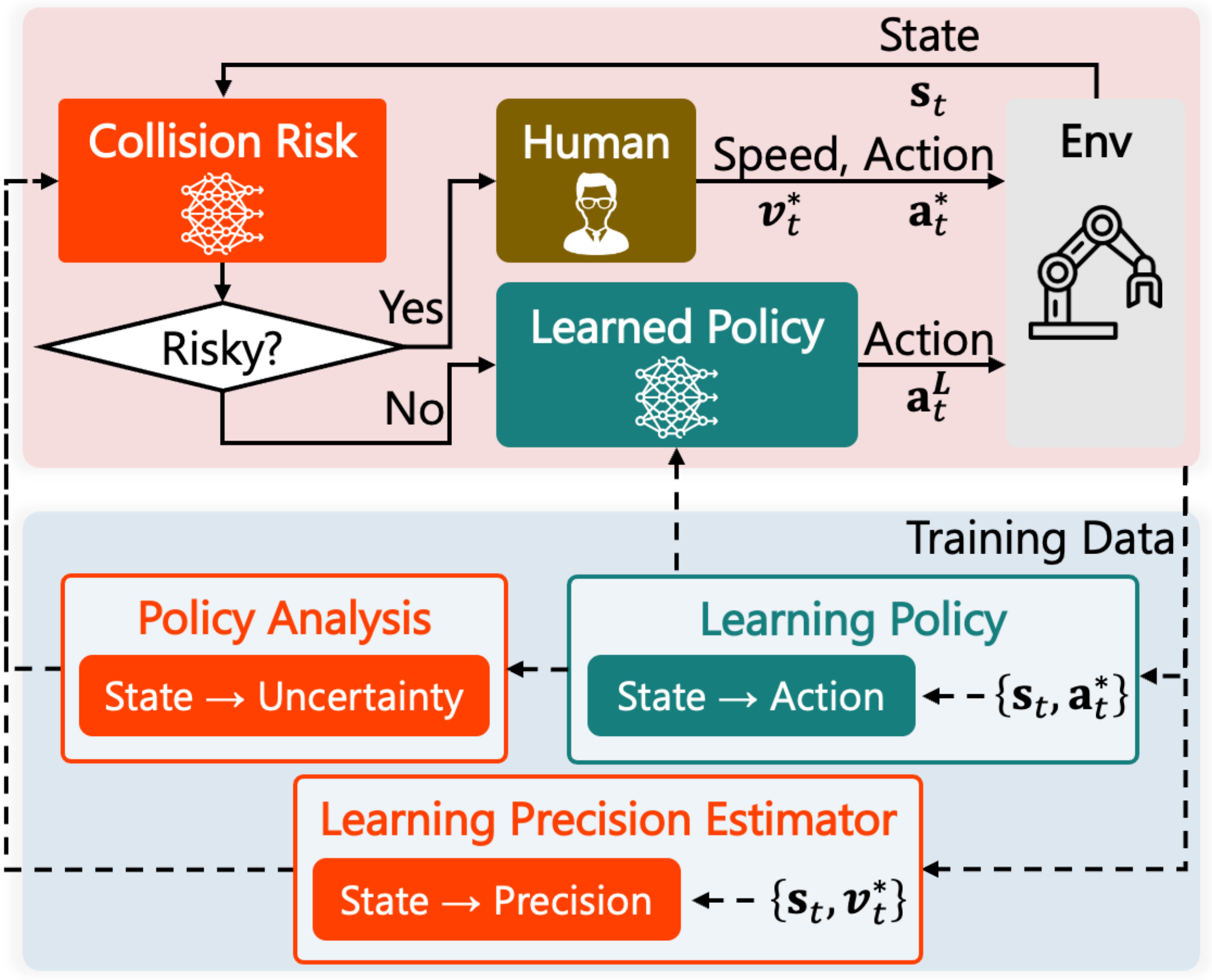}
    \caption{
        Overview of DPIIL: 
        (top): While a robot is executing a task with its policy, if $\mathbf{s}_t$ is too risky, a human controls it until the risk is sufficiently lowered. 
        (bottom): Policy and precision estimator are iteratively learned from training data collected through interactions. 
        Collision risk is computed with analyzed uncertainty of learned policy and estimated precision.
}
    \label{fig:overview}
\end{figure}

This section describes DPIIL's algorithmic flow. As shown in \fref{fig:overview}, the robot's policy is learned by iterating two phases: 
\begin{enumerate*}
    \item collecting training datasets through human-robot online interaction with collision risk estimation (\fref{fig:overview}, top), and 
    \item learning the robot's policy and the precision estimator using collected training datasets (\fref{fig:overview}, bottom).
\end{enumerate*}

Specifically, an initial dataset, $\mathcal{D} = \{\mathbf{a}^*_t,\mathbf{s}_t, v^*_t\}_{t=1}^{T}$, is only collected by $\pi_{\theta^*}$. The initial parameters of policy $\theta^{L}_0$ and precision estimator $\lambda_{0}$ are obtained by optimizing \eref{eq:il:obj} and \eref{eq:pros:speed:obj} on $\mathcal{D}$. Under this initialization, a training dataset is collected with an underlying interaction design (\sref{sec:proposal:interaction}) for $K$ iterations. At each $k$~th iteration, $\theta^{L}_{k-1}$ is used for the robot policy, and the states that are performed in the expert mode and the expert's actions and speeds are collected: $\mathcal{D}_{k} = \{\mathbf{a}^*_t,\mathbf{s}_t, v^*_t \mid g(\mathbf{s}_t;\chi)=1\}_{t=1}^{T}$.
These collected data are added to dataset $\mathcal{D}$. After each $k$~th iteration, the parameters of learned policy $\theta^{L}_{k}$ and precision estimator $\lambda_{k}$ are optimized using equations \eref{eq:il:obj} and \eref{eq:pros:speed:obj} on accumulated dataset $\mathcal{D}$.
A summary of DPIIL is shown in \aref{algorithm:DPIIL}.


\section{Simulation}
\label{sec:sim}
\begin{figure*}[t!]
    \centering
    \includegraphics[width=1.0\hsize]{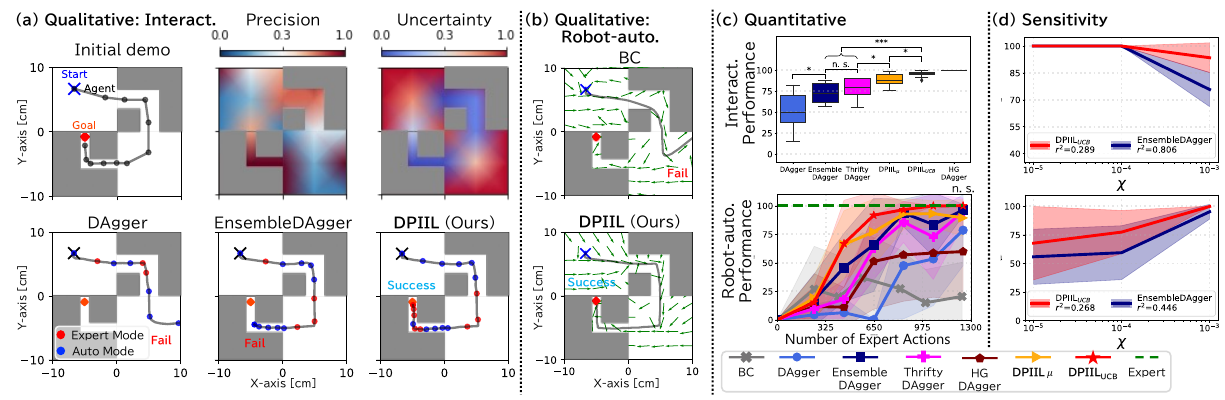}
    \caption{
    Aperture-passing simulation:
    \textbf{(a)}: 
    Uncertainty and precision results across state space are obtained using a policy and a precision estimator learned from initial demonstration dataset.
    Both measurements are normalized to clarify variations across states.
    Based on these indicators, interactive trajectories of IIL algorithms (DAgger, EnsembleDAgger, DPIIL (Ours)) are compared.
    \textbf{(b)}: 
    Comparison of the 2D vector fields of the policies learned by BC and DPIIL (ours) and their execution trajectories.
    \textbf{(c)}: 
    Averaged performance of interactive (top) and robot-autonomous (bottom) are evaluated by repeating each experiment ten times with random seeds.
    (top): Interactive performance is measured as a box plot of average success probability during training phases across entire trials of each IIL approach. Significant differences by t-test are observed between proposed method and a baseline ($*:p < 5e{-2}, ***:p < 5e{-4}$).
    (bottom): Comparing robot-autonomous performance for number of expert actions used to train by conducting 100 test episodes of each learned policy.
    The t-test results showed no significant difference between our method and other risk-aware IIL methods (EnsembleDAgger and ThriftyDAgger), but a significant difference ($p < 5e{-2}$) with HG-DAgger.
    \textbf{(d)}:
    Interactive and robot-autonomous performances are measured as $\chi$ values fixed at $\chi \in [10^{-5}, 10^{-3}]$ for each experiment; square of correlation coefficient $r^2$ \cite{hamby1994param_sensitivity} between hyperparameter $\chi$ and each performance is measured as sensitivity indicator.
}
    \label{fig:wall:env}
\end{figure*}

In this section, we validated whether our proposed method can effectively achieve an automation performance of a robot more safely than the prior algorithms in the following two simulation domains:
\begin{enumerate*}
    \item an aperture-passing task (\fref{fig:wall:env}) and 
    \item a ring-threading task with a 6-DOF UR5e robot (\fref{fig:peg:env}).
\end{enumerate*}

\textbf{Evaluation Metrics:}
The DPIIL performance is considered during the training and deployment test phases.
For the former, \textit{the interactive performance} was evaluated as the probability of the task's success over all the training episodes of the IIL approaches. 
For the latter, \textit{the robot-autonomous performance} was evaluated as the probability of the task's successful deployment of the learned policy after training without expert assistance.
Both metrics were assessed in both simulations (\sref{sec:sim:wall},\sref{sec:sim:peg}) and real-robot experiments (\sref{sec:exp}).

\textbf{Comparison Methods:}
In this evaluation, we compared our methods ($\text{DPIIL}_{\text{UCB}}$ and $\text{DPIIL}_{\mu}$) as a baseline to the following other imitation learning methods:
\begin{itemize}
    \item Behavior Cloning (BC)\cite{bain1995framework}: A conventional imitation learning that learns a policy without any interactions;
    \item Dataset Aggregation (DAgger)\cite{ross2011dagger}: a conventional IIL that randomly requests human intervention;
    \item EnsembleDAgger \cite{menda2019ensembledagger}: A state-of-the-art IIL that only uses policy decision uncertainty $\sigma_{\theta^{L}}^2$ as $\textrm{Risk}(\mathbf{s}_t)$.
    \item ThriftyDAgger \cite{hoque2021thriftydagger}: A state-of-the-art IIL where a precision estimator is learned through collision experiences.
    \item HG-DAgger \cite{kelly2019hgdagger}: A state-of-the-art IIL where an algorithmic expert decides when to intervene or not.
\end{itemize}
Our evaluation assumes an example problem where the ratio of states assigned as risky is sufficient and fair across risk-aware approaches (EnsembleDAgger, ThriftyDAgger, and DPIIL). To achieve this, we set the threshold $\chi$ for each method at the value of approximately the top $20\%$ of the estimated risk in the training dataset, similar to previous works \cite{zhang2016safedagger, Hoque2021lazydagger, hoque2021thriftydagger}. Its sensitivity is analyzed in \sref{sec:sim:wall}.

\textbf{Demonstration Setting:}
Initially, speed transformation function $f_v$ is defined by the Euclidean norm of the difference in position-related states $\mathbf{s}^{pos}_t \in \mathbf{s}_t$, which are generally included as the state space of robotic tasks (\eg positions of agent center or end effector): $f_v(\mathbf{s}_{t},\mathbf{s}_{t+1}) = \|\mathbf{s}^{pos}_{t+1}-\mathbf{s}^{pos}_{t}\|^2_2$.
Under this initial setting, demonstrations are provided by an algorithmic expert, especially where a human-like risk-sensitive movement \cite{nagengast2011risk} is implemented as shown in \fref{fig:wall:env}(a) and \fref{fig:peg:env}(a). 
Such movement is simulated by specifying agent's action for each state: fast in open areas (\eg far from walls), and slow in small clearance areas (\eg aperture traversal), while injecting state-dependent Gaussian noise \cite{harris1998signal} as described in \sref{sec:IV:A}. 
For an algorithmic expert in HG-DAgger, the timing of the intervention is also specified to prevent failure during interactions in \sref{sec:sim:wall}.

\subsection{Aperture-passing Simulation}
\label{sec:sim:wall}
An aperture-passing task involving multiple narrow apertures was initially performed in the OpenAI gym \cite{brockman2016gym} environment (\fref{fig:wall:env}(a)). 
In this experiment, interactive and robot-autonomous performances are evaluated in a challenging environment that includes states where such physical contacts are likely to occur as passing through narrow apertures, although no contacts are allowed for task success.

\subsubsection{Task Setting}
The task goal is to move the agent (black circles with a $0.25~\mathrm{cm}$ radius) clock-wise from the starting position through the apertures (each of which has a width of $3.0~\mathrm{cm}$ and $1.5~\mathrm{cm}$ sequentially) to the goal without colliding with the walls (gray).
The system state and action are the agent's position (\eg x, y-axis coordinates) and velocity (\eg x, y-axis).
The initial state is deviated by additive uniform noise $\epsilon_{\mathbf{s}_{0}}\sim\U(- 2~\mathrm{cm}, 2~\mathrm{cm})$.

\subsubsection{Learning Setting}
\label{sec:sim:wall:setup}
Under these experimental parameters, we collected three initial demonstration trajectories (248 state-action pairs) by the expert policy for all the comparisons.
This dataset is used to optimize initial learned policy $\pi_{\theta^{L}_0}$ and precision estimator $\text{Pre}_{\lambda_{0}}$.
For DPIIL and each IIL comparison method, an interactive demonstration is performed where the control mode switches between the auto and expert mode, only collecting state-action pairs that the expert controlled (\ie expert mode).
After collecting $200$ state-action pairs, the policy and precision estimator were updated
on the accumulated dataset.
If the agent collides with a wall, fails to reach the goal position within the time limit ($200$ steps), or moves beyond the task space, it is considered a failure.
This process is denoted as one $k$ iteration in \aref{algorithm:DPIIL} and is repeated $K=5$ times in this experiment.
For BC, demonstration datasets are additionally provided by expert policy only until the number of expert actions is roughly equivalent to the other IIL algorithms.

\subsubsection{Results}\label{sec:sim:wall:result}
The results are shown in \fref{fig:wall:env}.

\textbf{Qualitative Analysis:}
In terms of interactive performance, the interactive trajectories of the IIL methods (DAgger, EnsembleDAgger, and DPIIL) are compared in (\fref{fig:wall:env}(a)).
In DAgger, the timing of an expert's intervention is randomly decided during interactive demonstrations. Even if the agent has drifted away from the demo trajectories, expert intervention may not be requested timely, leading to failures (\eg leaving the task space).
EnsembleDAgger requests expert intervention when the uncertainty of the policy decision is high due to a lack of demo data.
Although this interaction design allows the robot to avoid drastic deviations from the demo trajectories, it cannot detect a collision risk in narrow states where slight deviations are unacceptable; expert intervention is not requested, resulting in failure (\eg collisions).
In contrast, our method (DPIIL) implicitly estimates the precision of the environment by observing the expert's demonstrations.
When the estimated precision is applied to detect the collision risk, expert interventions are encouraged in narrow states, resulting in successful interactions that avoid collisions.

In terms of robot-autonomous performance, learned policies of BC and DPIIL are compared in (\fref{fig:wall:env}(b)). In BC, the policy learned only near the initial trajectories, accumulating errors and failing execution.
In contrast, DPIIL can train the policy that recovers to the initial trajectory through interaction, resulting in successful execution.

\begin{figure*}[t!]
    \centering
    \includegraphics[width=1.0\hsize]{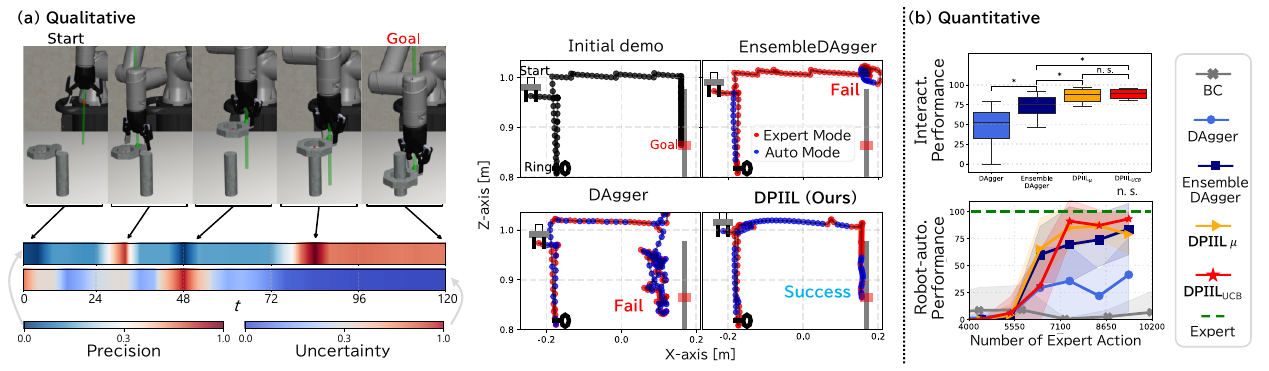}
    \caption{
    Ring-threading simulation:
    \textbf{(a)}: 
    Algorithmic expert's demonstration includes two high-precision phases as a robot reaches to grasp a ring and inserts it into a peg.
    Precision and uncertainty results were obtained by analyzing an initial demo using a precision estimator and a policy learned on the initial demo dataset.
    Based on this expert, interactive trajectories of IIL algorithms (DAgger, EnsembleDAgger, DPIIL (Ours)) were compared.
    \textbf{(b)}:
    Averaged performances of interactive (top) and robot-autonomous (bottom) were evaluated by repeating each experiment ten times with random seeds. Other details are identical as previous analysis (\fref{fig:wall:env}).
    }
    \label{fig:peg:env}
\end{figure*}

\textbf{Quantitative Analysis:}
We compared our methods with other baseline schemes regarding the interactive and robot-autonomous performances (\fref{fig:wall:env}(c)).
In terms of interaction performance, DAgger had poor performance ($52\%$) since its robot cannot be aware of any risks during the learned-policy execution. Although EnsembleDAgger has better performance ($73\%$) by considering the uncertainty of policy decisions and promoting expert intervention in highly uncertain states, it has next poor performance since it does not ask experts to intervene in states where collisions may occur, as predicted by our qualitative analysis. Despite utilizing precision estimation, ThriftyDAgger performs ($78\%$) similarly to EnsembleDAgger since it requires sufficient collision experience to estimate precision properly. In comparison, both our methods ($\text{DPIIL}_{\mu}$ and $\text{DPIIL}_{\text{UCB}}$) had significantly better performances ($89\%$ and $96\%$) than the others by using precision estimation without the collision experience, nearing the performance of an oracle (HG-DAgger) where an algorithmic expert decides when to intervene optimally.

In terms of robot-autonomous performance, BC performed poorly ($21\%$) as predicted by our qualitative analysis.
HG-DAgger monotonically increases the performance of the learned policy, but its performance is the worst ($60\%$) among the IIL methods. This is because the conservative expert repeatedly intervenes in a certain area and cannot generalize to a wider range of states.
The next worst IIL method is DAgger ($79\%$), since if the robot fails the task during the interactive demonstrations, it won't be able to continue training on the rest of the task progress, reducing the efficacy of interactive learning.
In contrast, risk-aware approaches can significantly improve performance (EnsembleDAgger: $96\%$, ThriftyDAgger: $95\%$, $\text{DPIIL}_{\mu}$: $89\%$). One of our methods ($\text{DPIIL}_{\text{UCB}}$) had the best performance ($100\%$) across all the iterations, suggesting that DPIIL increases interaction safety and ensures efficiency.

\textbf{Sensitivity Analysis of $\chi$:}
We analyzed and compared hyperparameter $\chi$'s sensitivity from the prior risk-aware approach (EnsembleDAgger) and our best method ($\text{DPIIL}_{\text{UCB}}$) (\fref{fig:wall:env}(d)).
EnsembleDAgger, which uses the uncertainty of the policy decision as risk, is sensitive to $\chi$, and the interactive and robot-autonomous performances are mutual trade-offs in a range of $\chi \in [10^{-4}, 10^{-3}]$.
In contrast, our method ($\text{DPIIL}_{\text{UCB}}$), which uses precision that is combined with uncertainty as a collision risk, is more robust to a wider range of $\chi$ in the interactive performance and has sufficient robot-autonomous performance at $\chi=10^{-3}$.

\subsection{Ring-threading Simulation}\label{sec:sim:peg}
To evaluate DPIIL's scalability, a second experiment was conducted for learning a ring-threading task with a 6-DOF UR5e robot in a Robosuite \cite{robosuite2020} environment (\fref{fig:peg:env}).
This task has two challenges that surpass an aperture-passing task: 
\begin{enumerate*}
\item various physical contact scenarios (\eg robot vs. object, object vs. object) can occur dynamically on high dimensional state-action space, and 
\item the ring and robot positions are randomly initialized.    
\end{enumerate*}

\begin{figure*}[ht]
    \centering
    \includegraphics[width=1.0\hsize]{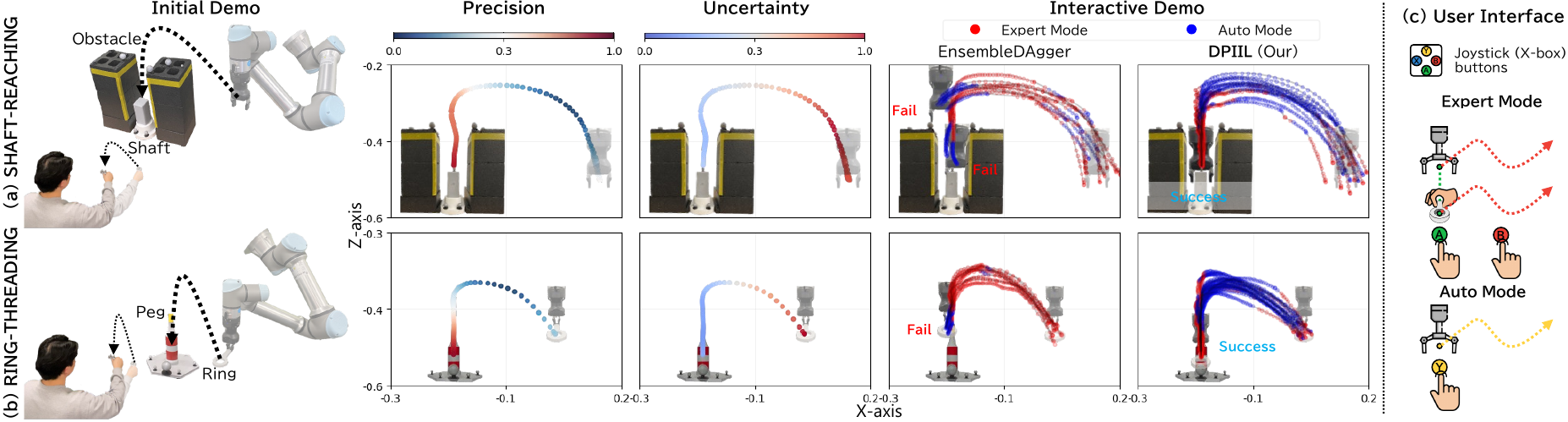}
    \caption{
    Real-robot experiments:
    Experiments were conducted for 6-DOF robotic arm (UR5e) assembly tasks with human experts: 
    \textbf{(a)} reaching a shaft by avoiding obstacles and 
    \textbf{(b)} threading a ring into a peg.
    Precision and uncertainty results were obtained by analyzing initial demonstration with a precision estimator and policy learned from initial dataset.
    Both measurements were normalized to visualize variations across states.
    An interactive demonstration of EnsembleDAgger and DPIIL (Ours) shows trajectories at interactive phase.
    \textbf{(c)}: Illustration of user interface using buttons on a joystick (X-box).
    In expert mode, pressing the "A" button synchronizes the position of the robot's end effector with that of the human-held ring. While the "B" button is pressed, the robot follows the movement of the ring. If the "B" button is released, the robot stops moving, and synchronization must be redone by pressing the "A" button again.
    In auto mode, while the "Y" button is pressed, the robot is moved by learned policy. Note, "Y" button is only set to ensure safety in verification evaluations, not as the requirement of our method (DPIIL).
    }
    \label{fig:real:env}
\end{figure*}

\subsubsection{Task Setting}
The goal is to grasp a ring with the random initial positions and insert it through a peg with a fixed position, regardless of the physical contact.
The dimension of the state is 51D, consisting of the robot's joint angles and the ring's position, and the action is 6D, specifying the end-effector translation (\eg x, y, z-axes), rotation (\eg y, z-axes), and gripper manipulation (\eg open or closed).
Further details can be seen at: \url{https://robosuite.ai/}.

\subsubsection{Learning Setting}
The procedure here is similar to \sref{sec:sim:wall:setup}, but due to the task's complexity, the amount of training data is increased by a factor of $10$. The number of initial demonstration trajectories collected by the expert policy is $30$ (\num[group-separator={,}]{4414} state-action pairs), and the number of state-action pairs collected by the expert mode in each iteration is \num[group-separator={,}]{1000}. Accordingly, the amount of training data for BC also increased. 
In addition, the time limit ($200$ steps) is this task's only failure condition for evaluating the performances under various physical contacts.

\subsubsection{Results}\label{sec:sim:peg:result}
The results can be seen in \fref{fig:peg:env}(a) and (b).

\textbf{Qualitative Analysis:}
We compared the interactive trajectories of the IIL methods (\fref{fig:peg:env}(a)).
As described in the previous qualitative analysis (\sref{sec:sim:wall:result}), the randomized intervention timing of DAgger may induce a robot to fall into a state where the task is infeasible even in the expert mode (\eg robotic arms getting tangled up), resulting in failure. 
Although the uncertainty-based interaction of EnsembleDAgger prevents vast deviations from the demo trajectory, it cannot detect precision to request an expert's intervention in high-precision areas, resulting in repeatedly failing to thread a ring due to slight deviations.
Contrarily, the DPIIL implicitly detects environmental precision from expert demonstrations to promote interventions in high-precision areas (\eg near a peg), resulting in successful interactive demonstrations.

\textbf{Quantitative Analysis:}
The overall results (\fref{fig:peg:env}(b)) show a similar trend to the previous task, although due to an increase in the task complexity, even $\text{DPIIL}_{\text{UCB}}$, which had the best performance in the previous results (\fref{fig:wall:env}(c)), required more than $10$ times the amount of training data to exceed the $90\%$ robot-autonomous performance ($93.3\%$). The other methods fail to even surpass $90\%$ despite the extra data.
As the amount of training data increased, the overall number of interactions also increased. Our methods ($\text{DPIIL}_{\text{UCB}}$ and $\text{DPIIL}_{\mu}$) still have significantly higher interactive performance, and the other methods have larger variance than the previous results (\fref{fig:wall:env}(c)) due to increased interactions.
These findings suggest that DPIIL can effectively address safety concerns in the interactive policy learning of clearance-limited tasks while ensuring efficiency.

\section{Real-Robot Experiments with Human Experts}
\label{sec:exp}
In this section, we verified the applicability of our method in various real-world scenarios (\fref{fig:real:env}) by conducting an experiment that trains the 6-DOF UR5e robot by human demonstrations of the following two assembly tasks:
\begin{enumerate}
    \item {a shaft-reaching task}: 
    We assessed the robot's skill to reach and grasp a shaft while avoiding fixed obstacles (\fref{fig:real:env}(a)). Successfully performing this task within the time limit ($150$ steps) is challenging since the environment is prone to physical contact (\eg robot vs. obstacles);
    \item {ring-threading task}: 
    We assessed the robot's skill of inserting a ring into a peg without bumping into another peg for the assembly (\fref{fig:real:env}(b)).
    This scenario is more complicated than the shaft-reach task since the clearance for inserting the ring is smaller (only $2~\mathrm{mm}$), requiring more precise control and a larger time limit ($200$ steps).
\end{enumerate}

\subsubsection{Task Setting}
The system state dimension is 12D, which consists of the robot's joint angles and the 3D coordinates of its arm and each task's target assembly part (\eg a shaft, a peg). The coordinate of each object (\eg a shaft, a peg, obstacles) are tracked by a motion capture system (OptiTrack Flex13) for detecting the collision to check task failure automatically.
An action is defined as the velocity of the robot arm in the x, y, and z-axes.
The initial robot end-effector position is deviated with additive uniform noise:
\begin{enumerate*}
    \item the shaft-reaching task: $\epsilon_{\mathbf{s}_{0}}\sim\U(- 0.05~\mathrm{m}, 0.05~\mathrm{m})$, and
    \item the ring-threading task: $\epsilon_{\mathbf{s}_{0}}\sim\U(- 0.02~\mathrm{m}, 0.02~\mathrm{m})$.
\end{enumerate*}

\subsubsection{Learning Setting}
In a similar procedure to \sref{sec:sim:wall:setup}, the human initially collects $5$ demonstration trajectories. The number of state-action pairs collected by the human expert in each iteration is $150$ for a shaft-reaching task and $200$ for a ring-threading task, and the number of iterations is $2$ ($K=2$). Accordingly, the amount of BC training data is roughly similar to the other IIL comparisons.

\textbf{Comparison Methods:}
In real-world evaluations, two approaches were compared with our methods as follows:
\begin{itemize}
    \item BC\cite{bain1995framework}: a conventional imitation learning;
    \item EnsembleDAgger \cite{menda2019ensembledagger}: a state-of-the-art risk-aware IIL.
\end{itemize}
Moreover, to ensure sufficient human analysis, more human actions are encouraged by setting threshold $\chi$ as $50\%$ of the overall training states that are classified as expert modes.

\textbf{Demonstration Setting:}
Demonstrations of each task were performed using a teleoperation system (\fref{fig:real:env}(c)) that synchronizes the robot's end effector with the position of a ring grasped by a human demonstrator. Thus, a robot follows a human hand's movements in a real-time manner. We used four human subjects with robotics experience. To obtain sufficient expert performance from them, the following curriculum was applied. Before starting each experiment, all subjects practiced teleoperating the robot by performing several task scenarios, ranging from wide clearance (\eg obstacle-free) to narrow clearance (\eg obstacle-present), until they became achieving success in each scenario consecutively. These interactions increased their understanding of environmental precision.
In addition, during task demonstrations, the subjects were informed of their remaining time by bells at every $1/3$ interval of the time limit.

\subsubsection{Results}

\begin{table*}[t!]
\caption{
    Real-robot experiments results:
    Performance of each learning model is mean and standard deviation of results of four subjects.
    Robot-autonomous performance of policies learned by each learning model was measured over ten test executions.
    Since BC is not IIL approach, we annotated it as $\mathrm{N/A}$ in interactive performances.
    The total number of interventions (mode switching from auto to expert) is measured as the factor of human stress.
    Our methods are significantly better than task results marked $^*$ (t-test, $p < 5e{-2}$).
    }
    \label{table:real:results}
\centering
\begin{tabular}{|p{2.5cm}|C{2cm}|C{2cm}|C{2cm}|C{2cm}|C{2cm}|C{2cm}|}
\hline
Learning & \multicolumn{2}{c|}{Interact. Perf. [$\%$]} & \multicolumn{2}{c|}{Robot-auto. Perf. [$\%$]} & \multicolumn{2}{c|}{Total \# of interventions} \\ \cline{2-7}
{Models}   & Shaft-reach. & Ring-thread. & Shaft-reach. & Ring-thread. & Shaft-reach. & Ring-thread. \\ \hline\hline
BC   & $\mathrm{N/A}$ & $\mathrm{N/A}$ & $0.0^* \pm 0.0$ & $0.0^* \pm 0.0$ & $\mathrm{N/A}$ & $\mathrm{N/A}$ \\ \hline
EnsembleDAgger   & $41.1^* \pm 19.4$ & $39.8^* \pm 19.1$ & $42.5^* \pm 30.3$ & $55.0^* \pm 26.9$ & $16.5 \pm 8.5$ & $20.3 \pm 4.6$    \\ \hline
$\textbf{DPIIL}_{\mu}$ (ours) & ${100.0} \pm 0.0$ & ${100.0} \pm 0.0$ & $82.5 \pm 13.0$ & $85.0 \pm 11.2$ & $12.25 \pm 4.76$ & $18.2 \pm 2.6$ \\ \hline
$\textbf{DPIIL}_\text{UCB}$ (ours) &${100.0} \pm 0.0$ & ${100.0} \pm 0.0$ & ${100.0} \pm 0.0$ & ${100.0} \pm 0.0$ & $10.5 \pm 2.7$ & $16.0 \pm 2.1$ \\ \hline
\end{tabular}
\end{table*}

The results are seen in \fref{fig:real:env} and \tref{table:real:results}.

\textbf{Qualitative Analysis:}
The interactive trajectories of the IIL methods are compared in \fref{fig:real:env}.
EnsembleDAgger uses the epistemic uncertainty of the policy as an intervention criterion and requests human intervention in highly uncertain areas (\eg randomly initialized starting position).
However, such policy uncertainty alone does not recognize the latent collision risks in the limited clearance areas due to obstacles.
Therefore, the robot is operated in the auto mode in narrow areas, and no human intervention is requested even when a collision is imminent, resulting in task failure.
In contrast, the proposed method (DPIIL) uses human demonstrations to capture environmental precision and incorporates it into an intervention criterion to recognize collision risks during the interaction phase.
Accordingly, the robot is operated in the expert mode during times of high collision risks (\eg near obstacles), thereby reducing their risk.

\textbf{Quantitative Analysis:}
The results (\tref{table:real:results}) show that BC has zero robot-autonomous performance in both the clearance-limited tasks.
This is because, as noted in a previous work \cite{ross2011dagger}, policies learned by BC easily lead a robot to deviate from human-demonstrated trajectories, and such deviations are not allowed in either task.
EnsembleDAgger outperformed BC, although it did not exceed $55\%$ in either one since frequent failures during interaction (less than $50\%$ of the interactive performance) make training on the task's later part insufficient.
Notably, our method (DPIIL) significantly improves both the interactive and robot-autonomous performances by at least $30\%$ compared to EnsembleDAgger in both tasks, without increasing the total number of interventions (\ie human stress).

\section{Discussion}
This paper presented DPIIL, a safe IIL algorithm that leverages demonstrator-perceived precision to mitigate collision risks during interactive policy learning. Our evaluations demonstrate that it can effectively learn clearance-limited tasks with significantly improved safety. Although, this paper assumes a demonstrator that has high sensitivity to precision, in practice, this situation may vary across individuals. For example, a demonstrator who emphasizes swiftly performing tasks at the expense of safety may operate the robot at high speeds even when high precision is required. Such human sensitivities can be captured as latent variables \cite{majumdar2017risk}, and our future work will explore how this changes DPIIL performances. In addition, we employed a robotic teleoperation system where a human movement is directly applied, exploiting human speed characteristics. For other teleoperation systems that rely on joystick controls, DPIIL can be extended by redefining speed with human decision-making times.

\bibliographystyle{IEEEtran}
\bibliography{reference}

\begin{thebibliography}{10}
\providecommand{\url}[1]{#1}
\csname url@samestyle\endcsname
\providecommand{\newblock}{\relax}
\providecommand{\bibinfo}[2]{#2}
\providecommand{\BIBentrySTDinterwordspacing}{\spaceskip=0pt\relax}
\providecommand{\BIBentryALTinterwordstretchfactor}{4}
\providecommand{\BIBentryALTinterwordspacing}{\spaceskip=\fontdimen2\font plus
\BIBentryALTinterwordstretchfactor\fontdimen3\font minus \fontdimen4\font\relax}
\providecommand{\BIBforeignlanguage}[2]{{%
\expandafter\ifx\csname l@#1\endcsname\relax
\typeout{** WARNING: IEEEtran.bst: No hyphenation pattern has been}%
\typeout{** loaded for the language `#1'. Using the pattern for}%
\typeout{** the default language instead.}%
\else
\language=\csname l@#1\endcsname
\fi
#2}}
\providecommand{\BIBdecl}{\relax}
\BIBdecl

\bibitem{osa2018algorithmic}
T.~Osa, J.~Pajarinen, G.~Neumann, J.~A. Bagnell, P.~Abbeel, and J.~Peters, ``An algorithmic perspective on imitation learning,'' \emph{Found. and Trends® in Robotics}, vol.~7, no. 1-2, pp. 1--179, 2018.

\bibitem{celemin2022interactive}
C.~Celemin, R.~P{\'e}rez-Dattari, E.~Chisari, G.~Franzese, L.~de~Souza~Rosa, R.~Prakash, Z.~Ajanovi{\'c}, M.~Ferraz, A.~Valada, J.~Kober \emph{et~al.}, ``Interactive imitation learning in robotics: A survey,'' \emph{Found. and Trends{\textregistered} in Robotics}, vol.~10, no. 1-2, pp. 1--197, 2022.

\bibitem{wickelgren1977speed}
W.~A. Wickelgren, ``Speed-accuracy tradeoff and information processing dynamics,'' \emph{Acta psychologica}, vol.~41, no.~1, pp. 67--85, 1977.

\bibitem{menda2019ensembledagger}
K.~Menda, K.~Driggs-Campbell, and M.~J. Kochenderfer, ``{EnsembleDAgger}: A {Bayesian} approach to safe imitation learning,'' in \emph{IEEE/RSJ Int. Conf. on Intelli. Robots and Sys.}, 2019, pp. 5041--5048.

\bibitem{zhang2016safedagger}
J.~Zhang and K.~Cho, ``Query-efficient imitation learning for end-to-end simulated driving,'' in \emph{Proceedings of the AAAI Conf. on Artificial Intelli.}, 2017, p. 2891–2897.

\bibitem{Hoque2021lazydagger}
R.~Hoque, A.~Balakrishna, C.~Putterman, M.~Luo, D.~S. Brown, D.~Seita, B.~Thananjeyan, E.~Novoseller, and K.~Goldberg, ``{LazyDAgger: Reducing Context Switching in Interactive Imitation Learning},'' in \emph{IEEE Int. Conf. on Autom. Sci. and Engineering}, 2021, pp. 502--509.

\bibitem{hoque2021thriftydagger}
R.~Hoque, A.~Balakrishna, E.~Novoseller, A.~Wilcox, D.~S. Brown, and K.~Goldberg, ``Thrifty{DA}gger: Budget-aware novelty and risk gating for interactive imitation learning,'' in \emph{Conf. on Robot Learning}, 2021, pp. 598--608.

\bibitem{laskey2017dart}
M.~Laskey, J.~Lee, R.~Fox, A.~Dragan, and K.~Goldberg, ``{DART}: Noise injection for robust imitation learning,'' in \emph{Conf. on Robot Learning}, 2017, pp. 143--156.

\bibitem{oh2023bdi}
H.~Oh, H.~Sasaki, B.~Michael, and T.~Matsubara, ``{Bayesian Disturbance Injection}: Robust imitation learning of flexible policies for robot manipulation,'' \emph{Neural Networks}, vol. 158, pp. 42--58, 2023.

\bibitem{ross2011dagger}
S.~Ross, G.~Gordon, and D.~Bagnell, ``A reduction of imitation learning and structured prediction to no-regret online learning,'' in \emph{Int. Conf. on Artificial Intelli. and Statistics}, 2011, pp. 627--635.

\bibitem{kelly2019hgdagger}
M.~Kelly, C.~Sidrane, K.~Driggs-Campbell, and M.~J. Kochenderfer, ``{HG-DAgger}: Interactive imitation learning with human experts,'' in \emph{IEEE Int. Conf. Robot. Autom.}, 2019, pp. 8077--8083.

\bibitem{hoque2023fleet}
R.~Hoque, L.~Y. Chen, S.~Sharma, K.~Dharmarajan, B.~Thananjeyan, P.~Abbeel, and K.~Goldberg, ``{Fleet-DAgger}: Interactive robot fleet learning with scalable human supervision,'' in \emph{Conf. on Robot Learning}, 2023, pp. 368--380.

\bibitem{nagengast2011risk}
A.~J. Nagengast, D.~A. Braun, and D.~M. Wolpert, ``Risk sensitivity in a motor task with speed-accuracy trade-off,'' \emph{Journal of neurophysiology}, vol. 105, no.~6, pp. 2668--2674, 2011.

\bibitem{lin2009speed}
H.-I. Lin and C.~G. Lee, ``Speed-accuracy optimization for skill learning,'' in \emph{IEEE Int. Conf. Robot. Autom.}, 2009, pp. 2506--2511.

\bibitem{murphy2011risky}
L.~Murphy and P.~Newman, ``Risky planning: Path planning over costmaps with a probabilistically bounded speed-accuracy tradeoff,'' in \emph{IEEE Int. Conf. Robot. Autom.}, 2011, pp. 3727--3732.

\bibitem{harris1998signal}
C.~M. Harris and D.~M. Wolpert, ``Signal-dependent noise determines motor planning,'' \emph{Nature}, vol. 394, no. 6695, pp. 780--784, 1998.

\bibitem{nix1994aleatoric}
D.~Nix and A.~Weigend, ``Estimating the mean and variance of the target probability distribution,'' in \emph{Proceedings of IEEE Int. Conf. on Neural Networks}, vol.~1, 1994, pp. 55--60.

\bibitem{hamby1994param_sensitivity}
D.~M. Hamby, ``A review of techniques for parameter sensitivity analysis of environmental models,'' \emph{Environmental monitoring and assessment}, vol.~32, no.~2, pp. 135--154, 1994.

\bibitem{bain1995framework}
M.~Bain and C.~Sammut, ``A framework for behavioural cloning.'' in \emph{Machine Intelli.}, 1995, pp. 103--129.

\bibitem{brockman2016gym}
G.~Brockman, V.~Cheung, L.~Pettersson, J.~Schneider, J.~Schulman, J.~Tang, and W.~Zaremba, ``Openai gym,'' in \emph{arXiv:1606.01540}, 2016.

\bibitem{robosuite2020}
Y.~Zhu, J.~Wong, A.~Mandlekar, and R.~Mart\'{i}n-Mart\'{i}n, ``robosuite: A modular simulation framework and benchmark for robot learning,'' in \emph{arXiv:2009.12293}, 2020.

\bibitem{majumdar2017risk}
A.~Majumdar, S.~Singh, A.~Mandlekar, and M.~Pavone, ``Risk-sensitive inverse reinforcement learning via coherent risk models.'' in \emph{Proceedings of Robotics: Sci. and Sys.}, 2017.

\end{thebibliography}

\end{document}